\pdfoutput=1

\documentclass[11pt]{article}

\usepackage{ACL2023}

\usepackage{times}
\usepackage{latexsym}

\usepackage[T1]{fontenc}

\usepackage[utf8]{inputenc}

\usepackage{microtype}

\usepackage{inconsolata}

\usepackage{color}
\usepackage{soul}
\usepackage{amsmath}
\usepackage{amsfonts}
\usepackage{amssymb}
\usepackage{pifont}
\usepackage{xcolor}
\usepackage{graphicx}
\usepackage{enumitem}
\usepackage{booktabs}
\usepackage{multirow}
\usepackage{array}
\usepackage{float}
\usepackage{arydshln}
\usepackage{adjustbox}
\usepackage{multicol}
\usepackage{mathabx}
\usepackage{soul}
\usepackage{cclicenses}
\usepackage[section]{placeins}
\usepackage[normalem]{ulem}
\usepackage{amsthm}

\usepackage[frozencache,cachedir=.]{minted}
\usepackage{tcolorbox}
\usepackage{etoolbox}
\newtcolorbox{codebox}{
    colback=lightgrey,
    colframe=grey,
    boxrule=0.5pt,
    arc=2mm,
    titlerule=0pt
}

\definecolor{lightgrey}{RGB}{245,245,245}
\definecolor{grey}{RGB}{155,155,155}

\usepackage[multiple]{footmisc}

\newcommand{\ensuretext}[1]{#1}

\definecolor{mdgreen}{rgb}{0.05,0.6,0.05}

\setlist[itemize]{itemsep=0.02cm,topsep=0.2cm}
\setlist[enumerate]{itemsep=0.02cm,topsep=0.2cm}

\title{\includegraphics[height=1.2\fontcharht\font`A]{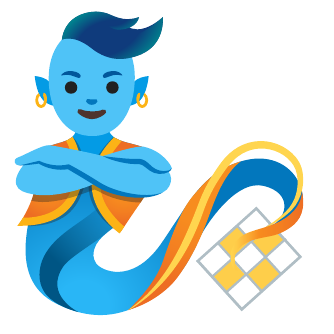} \textsc{TabGenie}: A Toolkit for Table-to-Text Generation}

\author{Zdeněk Kasner\textsuperscript{1}\quad Ekaterina Garanina\textsuperscript{1,2}\quad Ondřej Plátek\textsuperscript{1}\quad Ondřej Dušek\textsuperscript{1} \\
  \textsuperscript{1}Charles University, Czechia \\
  \textsuperscript{2}University of Groningen, The Netherlands \\
  \texttt{\{kasner,oplatek,odusek\}@ufal.mff.cuni.cz}\\\texttt{e.garanina@student.rug.nl}
}

\begin{document}
\maketitle
\begin{abstract}
Heterogenity of data-to-text generation datasets limits the research on data-to-text generation systems. We present \textsc{TabGenie} -- a toolkit which enables researchers to explore, preprocess, and analyze a variety of data-to-text generation datasets through the unified framework of \textit{table-to-text generation}. In \textsc{TabGenie}, all the inputs are represented as tables with associated metadata. The tables can be explored through the web interface, which also provides an interactive mode for debugging table-to-text generation, facilitates side-by-side comparison of generated system outputs, and allows easy exports for manual analysis. Furthermore, \textsc{TabGenie} is equipped with command line processing tools and Python bindings for unified dataset loading and processing. We release \textsc{TabGenie} as a PyPI package\footnote{\url{https://pypi.org/project/tabgenie/}} and provide its open-source code and a live demo at \url{https://github.com/kasnerz/tabgenie}.\footnote{Video: \url{https://youtu.be/iUC3NCGoFRg}}
\end{abstract}

\section{Introduction}
Building and evaluating data-to-text (D2T) generation systems \cite{gatt2018survey,sharma2022innovations} requires understanding the data and observing system behavior. It is, however, not trivial to interact with the large volume of D2T generation datasets that have emerged in the last years (see Table~\ref{tab:datasets}). 
Although research on D2T generation benefits from platforms providing unified interfaces, such as HuggingFace Datasets \cite{lhoest2021datasets} or the GEM benchmark \cite{gehrmann2021gem}, these platforms still leave the majority of the data processing load on the user.

A key component missing from current D2T tools is the possibility to visualize the input data and generated outputs. Visualization plays an important role in examining and evaluating scientific data \cite{Kehrer2013VisualizationAV} and can help D2T generation researchers to make more informed design choices. A suitable interface can also encourage researchers to step away from unreliable automatic metrics \cite{gehrmann2022repairing} and focus on manual error analysis \cite{van_miltenburg_underreporting_2021,van_miltenburg_barriers_2023}.

\begin{figure}[t]
  \centering
  \includegraphics[width=\columnwidth]{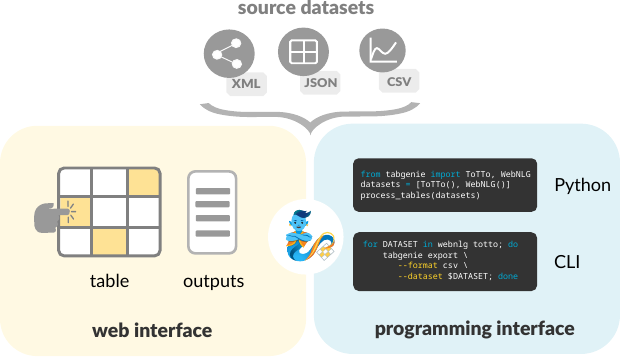}
  \caption{\textsc{TabGenie} provides a way to handle various data-to-text generation datasets through a unified web and programming interface. The \textit{web interface} enables interactive exploration and analysis of datasets and model outputs, while the \textit{programming interface} provides unified data loaders and structures.}\label{fig:teaser}
\end{figure}

Along with that, demands for a \textit{unified input data format} have recently been raised with multi-task training for large language models (LLMs) \citep[\textit{inter alia}]{Sanh2021MultitaskPT,scao2022bloom,Ouyang2022TrainingLM}. Some works have used simple data linearization techniques for converting structured data to a textual format, in order to align it with the format used for other tasks \cite{UnifiedSKG,tang2022mvp}. However,  linearizations are using custom preprocessing code, leading to discrepancies between individual works.

In this paper, we present \textsc{TabGenie} -- a multi-purpose toolkit for interacting with D2T generation datasets and systems designed to fill these gaps. On a high level, the toolkit consists of (a) an interactive web interface, (b) a set of command-line processing tools, and (c) a set of Python bindings (see Figure~\ref{fig:teaser}).

\begin{table*}[t]
	\centering\small
	\begin{tabular}{@{}lllrrrl@{}}
		\toprule
		\multirow{2}{*}{\textbf{Dataset}} & \multirow{2}{*}{\textbf{Source}} & \multirow{2}{*}{\textbf{Data Type}}  & \multicolumn{3}{c}{\textbf{Number of examples}} & \multirow{2}{*}{\textbf{License}} \\  \cmidrule(lr){4-6}
		                     &                                     &               & \textbf{train} & \textbf{dev} & \textbf{test} &               \\ \midrule 
		CACAPO               & \citet{van2020cacapo}               & Key-value     & 15,290         & 1,831        & 3,028         &  CC BY           \\
		DART$^\dagger$              & \citet{nan2021dart}                 & Graph & 62,659         & 2,768        & 5,097         & MIT           \\
		E2E$^\dagger$                  & \citet{duvsek2019semantic}          & Key-value     & 33,525         & 1,484        & 1,847         &   CC BY-SA         \\
		EventNarrative         & \citet{colas2021eventnarrative}     & Graph & 179,544        & 22,442       & 22,442        &  CC BY           \\
		HiTab                 & \citet{cheng2021hitab}              & Table w/hl & 7,417          & 1,671        & 1,584         & C-UDA         \\
		Chart-To-Text         & \citet{kantharaj2022chart}          & Chart & 24,368         & 5,221        & 5,222         & GNU GPL       \\
		Logic2Text            & \citet{chen2020logic2text}          & Table w/hl + Logic & 8,566          & 1,095        & 1,092         & MIT           \\
		LogicNLG              & \citet{chen2020logical}             & Table & 28,450         & 4,260        & 4,305         & MIT           \\
		NumericNLG            & \citet{suadaa-etal-2021-towards}    & Table & 1,084          & 136          & 135           &   CC BY-SA         \\
		SciGen                &  \citet{Moosavi2021LearningTR}                                   & Table & 13,607         & 3,452        & 492           &  CC BY-NC-SA       \\
		SportSett:Basketball$^\dagger$ & \citet{thomson-etal-2020-sportsett} & Table & 3,690          & 1,230        & 1,230         & MIT           \\
		ToTTo$^\dagger$                & \citet{parikh2020totto}             & Table w/hl & 121,153        & 7,700        & 7,700         &   CC BY-SA         \\
		WebNLG$^\dagger$               & \citet{ferreira20202020}            & Graph & 35,425         & 1,666        & 1,778         &  CC BY-NC         \\
		WikiBio$^\dagger$              & \citet{lebret2016neural}            & Key-value     & 582,659        & 72,831       & 72,831        &   CC BY-SA         \\
		WikiSQL$^\dagger$              & \citet{zhongSeq2SQL2017}                                    & Table + SQL   & 56,355         & 8,421        & 15,878        & BSD  \\
		WikiTableText         & \citet{bao2018table}            & Key-value     & 10,000         & 1,318        & 2,000         &  CC BY           \\
		\bottomrule
	\end{tabular}
	\caption{The list of datasets included in \textsc{TabGenie}. Glossary of data types: \textit{Key-value}: key-value pairs, \textit{Graph}: subject-predicate-object triples, \textit{Table}: tabular data (\textit{w/hl}: with highlighted cells), \textit{Chart}: chart data, \textit{Logic / SQL}: strings with logical expressions / SQL queries.  The datasets marked with $\dagger$ were already present on Huggingface Datasets. We uploaded the rest of the datasets to our namespace: \texttt{\url{https://huggingface.co/kasnerz}}.}
	\label{tab:datasets}.
\end{table*}

The cornerstone of \textsc{TabGenie} is a \textbf{unified data representation}. Each input represented is as a matrix of $m$ columns and $n$ rows consisting of individual cells accompanied with metadata (see §\ref{sec:data}). Building upon this representation, \textsc{TabGenie} then provides multiple features for unified workflows with table-to-text datasets, including:
\begin{enumerate}
    \item visualizing individual dataset examples in the tabular format (§\ref{sec:exploration}),
    \item interacting with table-to-text generation systems in real-time (§\ref{sec:interactive}),
    \item comparing generated system outputs (§\ref{sec:interactive}),
    \item loading and preprocessing data for downstream tasks (§\ref{sec:python}),
    \item exporting examples and generating spreadsheets for manual error analysis (§\ref{sec:cli}).
\end{enumerate}

In §\ref{sec:casestudies}, we present examples of practical use-cases of \textsc{TabGenie} in D2T generation research.

\section{Data}
\label{sec:data}
We currently include 16 datasets listed in Table~\ref{tab:datasets} in \textsc{TabGenie}, covering many subtasks of D2T generation. All the datasets are available under a permissive open-source license. 

\begin{figure*}[t]
  \centering
  \setlength{\fboxsep}{0pt}\fcolorbox{gray!20}{white}{\includegraphics[width=1.0\textwidth]{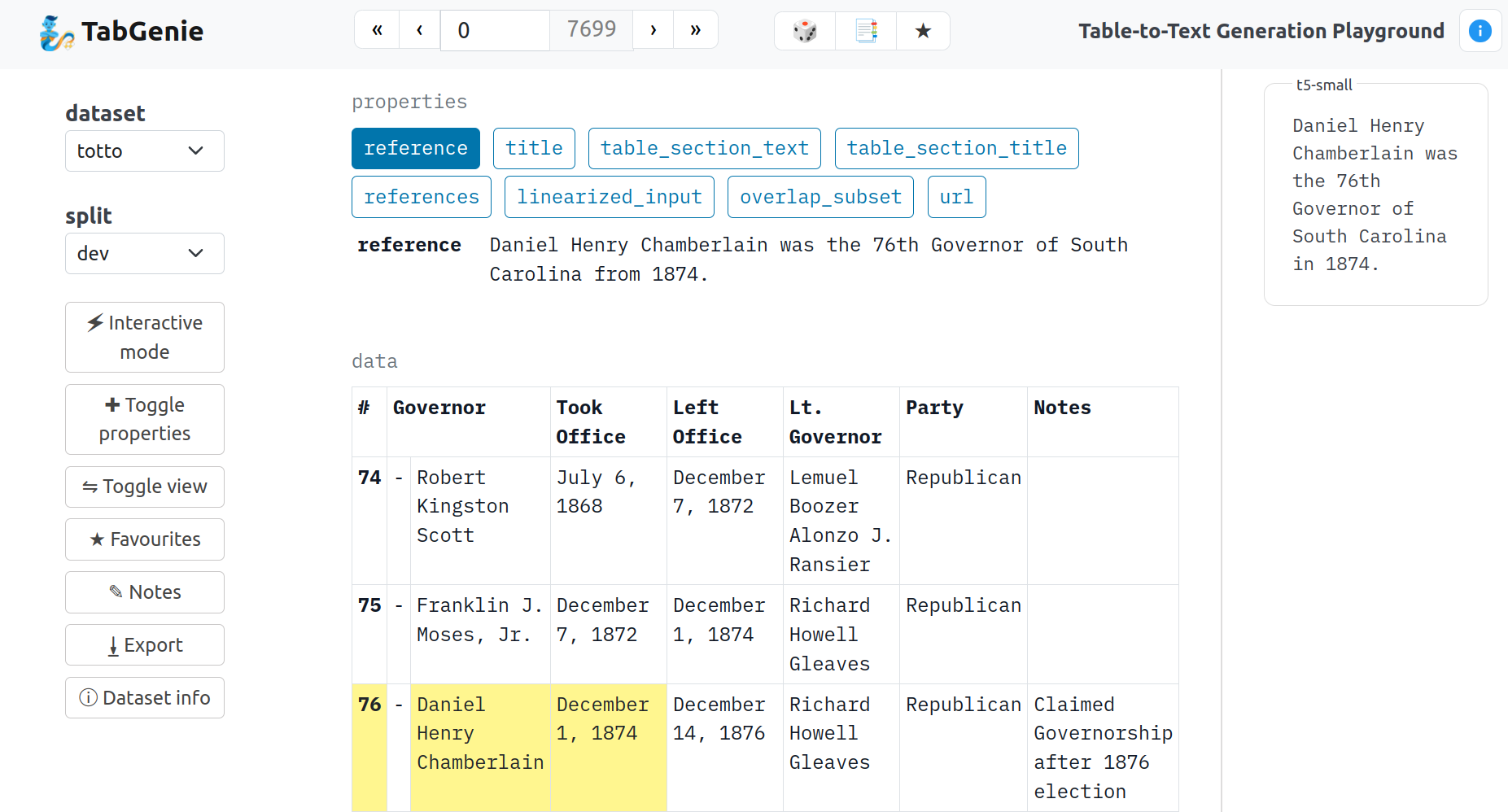}}
  \caption{The web interface of \textsc{TabGenie}. The \textbf{left panel} and the \textbf{navigation bar} contains user controls; the \textbf{center panel} shows table properties and table content; the \textbf{right panel} contains system outputs.}
  \label{fig:web}
\end{figure*}

\subsection{Data Format}
The inputs in D2T generation datasets may not consist only of tables, but also of e.g.\ graphs or key-value pairs. However, we noticed that in many cases, converting these formats to tables requires only minimal changes to the data structure while allowing a unified data representation and visualization. This conversion narrows down the task of D2T generation as the task of generating description for a tabular data, i.e. table-to-text generation \cite{parikh2020totto, liu2022plog, gong2020tablegpt}.

In our definition, a \textit{table} is a two-dimensional matrix with $m$ columns and $n$ rows, which together define a grid of $m \times n$ cells. Each cell contains a (possibly empty) text string. A continuous sequence of cells $\{c_{i}, \ldots, c_{i+k}\}$ from the same row or column may be merged, in which case the values of $\{c_{i+1},\ldots,c_{i+k}\}$ are linked to the value of $c_{i}$.  A cell may be optionally marked as a \textit{heading}, which is represented as an additional property of the cell.\footnote{The headings are typically located in the first row or column, but may also span multiple rows or columns and may not be adjacent.} To better accommodate the format of datasets such as ToTTo \cite{parikh2020totto} or HiTab \cite{cheng2021hitab}, we also allow individual cells to be \textit{highlighted}. Highlighted cells are assumed to be preselected for generating the output description.

The tables may be accompanied with an additional set of properties (see Figure~\ref{fig:web}) -- an example of such a property is a \textit{``title''} of the table in WikiBio \cite{lebret2016neural} or a \textit{``category''} in WebNLG \cite{gardent2017webnlg}. We represent properties as key-value pairs alongside the table. The properties may be used for generating the table description.

\subsection{Data Transformation}
We aim to present the data as true to the original format as possible and only make some minor changes for datasets which do not immediately adhere to the tabular format:

\begin{itemize}
  \item For graph-to-text datasets, we format each triple as a row, using three columns labeled \textit{subject}, \textit{predicate}, and \textit{object}.
  \item For key-value datasets, we use two columns with keys in the first column as row headings.
  \item For SportSett:Basketball \cite{thomson-etal-2020-sportsett}, we merge the \textit{box score} and \textit{line score} tables and add appropriate headings where necessary.
\end{itemize}

\subsection{Data Loading}
To ease the data distribution, we load all the datasets using the Huggingface \texttt{datasets} package \cite{lhoest2021datasets}, which comes equipped with a data downloader. Out of 16 datasets we are using, 7 were already available in Huggingface datasets, either through the GEM benchmark \cite{gehrmann2021gem} or other sources. We publicly added the 9 remaining datasets (see Table~\ref{tab:datasets}).

\textsc{TabGenie} also supports adding custom data loaders. Creating a data loader consists of simple sub-classing the data loader class and overriding a single method for processing individual entries, allowing anyone to add their custom dataset.

\section{Web Interface}
\label{sec:web}

\textsc{TabGenie} offers a user-friendly way to interact with table-to-text generation datasets through the \textit{web interface}. The interface can be rendered using a local server (cf. §\ref{sec:cli}) and can be viewed in any modern web browser. The interface features a simple, single-page layout, which contains a navigation bar and three panels containing user controls, input data, and system outputs (see Figure \ref{fig:web}). Although the interface primarily aims at researchers, it can be also used by non-expert users.

\subsection{Content Exploration}
\label{sec:exploration}
The input data in \textsc{TabGenie} is rendered as HTML tables, providing better visualizations than existing data viewers, especially in the case of large and hierarchical tables.\footnote{Compare, e.g., with the ToTTo dataset in Huggingface Datasets for which the table is provided in a single field called \textit{``table''}: \url{https://huggingface.co/datasets/totto}} In the web interface, users can navigate through individual examples in the dataset sequentially, access an example using its index, or go to a random example. The users can add notes to examples and mark examples as favorites for accessing them later. The interface also shows the information about the dataset (such as its description, version, homepage, and license) and provides an option to export the individual examples (see §\ref{sec:cli}).

\subsection{Interactive Mode}
\label{sec:interactive}
\textsc{TabGenie} offers an \textit{interactive mode} for generating an output for a particular example on-the-fly. The user can highlight different cells, edit cell contents, and edit parameters of the downstream processor. For example, the user can prompt a LLM for table-to-text generation and observe how it behaves while changing the prompt.

The contents of a table are processed by a processing \textit{pipeline}. This pipeline takes table contents and properties as input, processes them with a sequence of modules, and outputs HTML code. The modules are custom Python programs which may be re-used across the pipelines.

\textsc{TabGenie} currently provides two basic pipelines: (1) calling a generative language model through an API with a custom prompt, and (2) generating graph visualizations of RDF triples. We describe the case-study for the model API pipeline in §\ref{sec:cs:prompting}. Users can easily add custom pipelines by following the instructions in the project repository.

\subsection{Pre-generated Outputs}
\label{sec:outputs}
In addition to interactive generation, \textsc{TabGenie} allows to visualize static pre-generated outputs. These are loaded in the JSONL\footnote{\url{https://jsonlines.org}} format from the specified directory and displayed similarly to the outputs from the interactive mode. Multiple outputs can be displayed alongside a specific example, allowing to compare outputs from multiple systems.

\section{Developer Tools}
\label{sec:developer}

\textsc{TabGenie} also provides a developer-friendly interface: Python bindings (§\ref{sec:python}) and a command-line interface (§\ref{sec:cli}). Both of these interfaces aim to simplify dataset preprocessing in downstream tasks. The key benefit of using \textsc{TabGenie} is that it provides streamlined access to data in a consistent format, removing the need for dataset-specific code for extracting information such as table properties, references, or individual cell values.

\subsection{Python Bindings}
\label{sec:python}
\textsc{TabGenie} can be integrated in other Python codebases to replace custom preprocessing code. With a \textit{single unified interface} for all the datasets, the \textsc{TabGenie} wrapper class allows to:
\begin{itemize}
    \item load a dataset from the Huggingface Datasets or from a local folder,
    \item access individual table cells and their properties,
    \item linearize tables using pre-defined or custom functions,
    \item prepare the Huggingface \texttt{Dataset} objects for downstream processing.
\end{itemize}
\textsc{TabGenie} can be installed as a Python package, making the integration simple and intuitive.
See §\ref{sec:cs:generation} for an example usage of the \textsc{TabGenie} Python interface.

\subsection{Command-line Tools}
\label{sec:cli}
\textsc{TabGenie} supports several basic commands via command line.

\paragraph{Run} The \texttt{tabgenie run} command launches the local web server, mimicking the behavior of \texttt{flask run}. Example usage:

\begin{codebox}
\begin{minted}[fontsize=\small]{bash}
tabgenie run --port=8890 --host="0.0.0.0"
\end{minted}
\end{codebox}

\paragraph{Export} The \texttt{tabgenie export} command enables batch exporting of the dataset. The supported formats are \texttt{xlsx}, \texttt{html}, \texttt{json}, \texttt{txt}, and \texttt{csv}. Except for \texttt{csv}, table properties can be exported along with the table content. Example usage:

\begin{codebox}
\begin{minted}[fontsize=\small]{bash}
tabgenie export --dataset "webnlg" \
  --split "dev" \
  --out_dir "export/datasets/webnlg" \
  --export_format "xlsx"
\end{minted}
\end{codebox}
\noindent Export can also be done in the web interface.

\paragraph{Spreadsheet} For error analysis, it is common to select $N$ random examples from the dataset along with the system outputs and manually annotate them with error categories (see~§\ref{sec:cs:analysis}). The \texttt{tabgenie sheet} command generates a suitable spreadsheet for this procedure. Example usage:

\begin{codebox}
\begin{minted}[fontsize=\small]{bash}
tabgenie sheet --dataset "webnlg" \
  --split "dev" \
  --in_file "out-t5-base.jsonl" \
  --out_file "analysis_webnlg.xlsx" \
  --count 50
  \end{minted}
  \end{codebox}

\section{Implementation}
\label{sec:architecture}

\textsc{TabGenie} runs with Python >=3.8 and requires only a few basic packages as dependencies. It can be installed as a stand-alone Python module from PyPI (\texttt{pip install tabgenie}) or from the project repository.

\paragraph{Backend} The web server is based on \texttt{Flask},\footnote{\url{https://pypi.org/project/Flask/}} a popular lightweight Python-based web framework. The server runs locally and can be configured with a YAML\footnote{\url{https://yaml.org}} configuration file. On startup, the server loads the data using the \texttt{datasets}\footnote{\url{https://pypi.org/project/datasets/}} package. To render web pages, the server uses the \texttt{tinyhtml}\footnote{\url{https://pypi.org/project/tinyhtml/}} package and Jinja\footnote{\url{https://jinja.palletsprojects.com/}} templating language. 

\paragraph{Frontend} The web frontend is built on HTML5, CSS, Bootstrap,\footnote{\url{https://getbootstrap.com/}} JavaScript, and jQuery.\footnote{\url{https://jquery.com}} We additionally use the D3.js\footnote{\url{https://d3js.org}} library for visualizing the structure of data in graph-to-text datasets. To keep the project simple, we do not use any other major external libraries.

\section{Case Studies}
\label{sec:casestudies}

In this section, we outline several recipes for using \textsc{TabGenie} in D2T generation research. The instructions and code samples for these tasks are available in the project repository.

\subsection{Table-To-Text Generation} 
\label{sec:cs:generation}
  \paragraph{Application} Finetuning a sequence-to-sequence language model for table-to-text generation in PyTorch \cite{paszke2019pytorch} using the Huggingface Transformers \cite{wolf2019huggingface} framework.

  \paragraph{Process} In a typical finetuning procedure using these frameworks, the user needs to prepare a \texttt{Dataset} object with tokenized input and output sequences. Using \textsc{TabGenie}, preprocessing a specific dataset is simplified to the following:

\begin{codebox}
\begin{minted}[fontsize=\fontsize{8pt}{8pt}]{python}
from transformers import AutoTokenizer
import tabgenie as tg

# instantiate a tokenizer
tokenizer = AutoTokenizer.from_pretrained(...)

# load the dataset
tg_dataset = tg.load_dataset(
    dataset_name="totto"
)

# preprocess the dataset
hf_dataset = tg_dataset.get_hf_dataset(
    split="train",
    tokenizer=tokenizer
)
\end{minted}
\end{codebox}
 
The function \texttt{get\_hf\_dataset()} linearizes the tables (the users may optionally provide their custom linearization function) and tokenizes the inputs and references.

For training a single model on multiple datasets in the multi-task learning setting \cite{UnifiedSKG}, the user may preprocess each dataset individually, prepending a dataset-specific task description to each example. The datasets may then be combined using the methods provided by the \texttt{datasets} package.

\paragraph{Demonstration} For running the baselines, we provide an example script, which can be applied to any \textsc{TabGenie} dataset and pre-trained sequence-to-sequence model from the \texttt{transformers} library. For multi-task learning, we provide an example of joint training on several datasets with custom linearization functions. We run the example scripts for several datasets and display the resulting generations in the application demo. Details on the fine-tuned models can be found in Appendix \ref{appendix:models}.

\subsection{Interactive Prompting}
\label{sec:cs:prompting}

  \paragraph{Application} Observing the impact of various inputs on the outputs of a LLM prompted for table-to-text generation.
  \paragraph{Process} The user customizes the provided \texttt{model\_api} pipeline to communicate with a LLM through an API. The API can communicate either with an external model (using e.g. OpenAI API\footnote{\url{https://openai.com/api/}}), or with a model running locally (using libraries such as FastAPI\footnote{\url{https://fastapi.tiangolo.com}}). The user then interacts with the model through \textsc{TabGenie} web interface, modifying the prompts, highlighted cells, and table content (see §\ref{sec:interactive}).
  \paragraph{Demonstration} We provide an interactive access to the instruction-tuned Tk-Instruct \texttt{def-pos-11b} LLM \cite{wang2022super} in the project live demo. The user can use the full range of possibilities included in the interactive mode, including customizing the prompt and the input data.\footnote{Note that using the model for the task of table-to-text generation is experimental and may not produce optimal outputs. The model should also not be used outside of demonstration purposes due to our limited computational resources.} The interface is shown in Appendix \ref{appendix:screenshots}.

\subsection{Error Analysis}
\label{sec:cs:analysis}

\paragraph{Application}  
Annotating error categories in the outputs from a table-to-text generation model.

\paragraph{Process} The user generates the system outputs (see §\ref{sec:cs:generation}) and saves the outputs for a particular dataset split in a JSONL format. Through the command-line interface, the user will then generate a XLSX file which can be imported in any suitable office software and distributed to annotators for performing error analysis. 

\paragraph{Demonstration} We provide instructions for generating the spreadsheet in the project documentation. See Appendix \ref{appendix:screenshots} for a preview of the spreadsheet format.

\section{Related Work}

\subsection{Data Loading and Processing}
As noted throughout the work, Huggingface Datasets \cite{lhoest2021datasets} is the primary competitor package for data loading and preprocessing. Our package serves as a wrapper on top of this framework, providing additional abstractions for D2T generation datasets.

DataLab \cite{xiao-etal-2022-datalab} is another platform for working with NLP datasets. Similarly to Huggingface Datasets, this platform has much broader focus than our package. Besides data access, it offers fine-grained data analysis and data manipulation tools. However, it has limited capabilities of visualizing the input data or interactive generation and at present, it does not cover the majority of datasets available in \textsc{TabGenie}.

PromptSource \cite{bach2022promptsource} is a framework for constructing prompts for generative language models using the Jinja templating language. It can be used both for developing new prompts and for using the prompts in downstream applications.

Several tools have been developed for comparing outputs of language generation systems (notably for machine translation) such as CompareMT \cite{neubig2019comparemt} or Appraise \cite{federmann2018appraise}, but the tools do not visualize the structured data.

\subsection{Interactive D2T Generation}

Until now, platforms for interactive D2T generation have been primarily limited to commercial platforms, such as Arria,\footnote{\url{https://www.arria.com}} Automated Insights,\footnote{\url{https://automatedinsights.com}} or Tableau Software\footnote{\url{https://www.tableau.com}} (formerly Narrative Science). These platforms focus on proprietary solutions for generating business insights and do not provide an interface for research datasets. \citet{dou-etal-2018-data2text} present Data2Text Studio, a platform which provides a set of developer tools for building custom D2T generation systems. The platform currently does not seem to be publicly available.

\subsection{Table-To-Text Generation}
Although pre-trained sequence-to-sequence models have been found to be effective for D2T generation \citep{kale-rastogi-2020-text, UnifiedSKG}, they have difficulties with handling the input structure, generation diversity, and logical reasoning. Multiple works have tried to address these issues. For a comprehensive review of the field, we point out the interested reader to the recent survey of \citet{sharma2022innovations}.

\section{Conclusion}
We presented \textsc{TabGenie}, a multifunctional software package for table-to-text generation. \textsc{TabGenie} bridges several gaps including visualizing input data, unified data access, and interactive table-to-text generation. As such, \textsc{TabGenie} provides a comprehensive set of tools poised to accelerate progress in the field of D2T generation.

\section*{Limitations}
For some D2T generation inputs, the tabular structure may be inappropriate. This involves hierarchical tree-based structures, bag-of-words, or multimodal inputs \cite{balakrishnan2019constrained,lin2019commongen,krishna2017visual}. Due to deployment issues, \textsc{TabGenie} also does not include large synthetic datasets \cite{agarwal2021knowledge,jin2020genwiki}. \textsc{TabGenie} is currently in early development stages, which is why it primarily targets the research community.

\section*{Ethical Impact}

The table-to-text generation datasets may contain various biases or factually incorrect outputs, which may be further reproduced by the table-to-text generation models. Although our software package is designed to help to examine and eliminate the biases and errors, we cannot guarantee the correctness of the processed outputs.

As \textsc{TabGenie} is an open-source software package with a permissive license, we do not control its downstream applications. We advocate using it for responsible research with the aim of improving natural language generation systems.

\bibliography{anthology,custom}
\bibliographystyle{acl_natbib}

\appendix

\section{Fine-tuned models}
\label{appendix:models}

For the demo purposes, we have fine-tuned the following models using our example scripts:
\begin{itemize}
\item \texttt{t5-small} for Chart-To-Text, LogicNLG, ToTTo, WikiTableText;
\item \texttt{t5-base} for DART, E2E, WebNLG;
\item \texttt{t5-base} in a prefix-based multi-task setup on E2E and WebNLG, using custom linearization functions.
\end{itemize}

All models (individual and multi-task) were fine-tuned using \texttt{transformers} library. The parameters are the following:

\begin{itemize}
\item Epochs: 30 or individual models and 15 for multi-task,
\item Patience: 5 epochs,
\item Batch size: 16,
\item Optimizer: AdamW,
\item Learning rate: 1e-4,
\item Weight decay: 0,
\item AdamW betas: 0.9, 0.999,
\item Maximum input length: 512,
\item Maximum output length: 512,
\item Generation beam size: 3.
\end{itemize}

\section{User Interface}
\label{appendix:screenshots}

\begin{figure*}[htb]
  \centering
  \setlength{\fboxsep}{0pt}\fcolorbox{gray!20}{white}{\includegraphics[width=\textwidth]{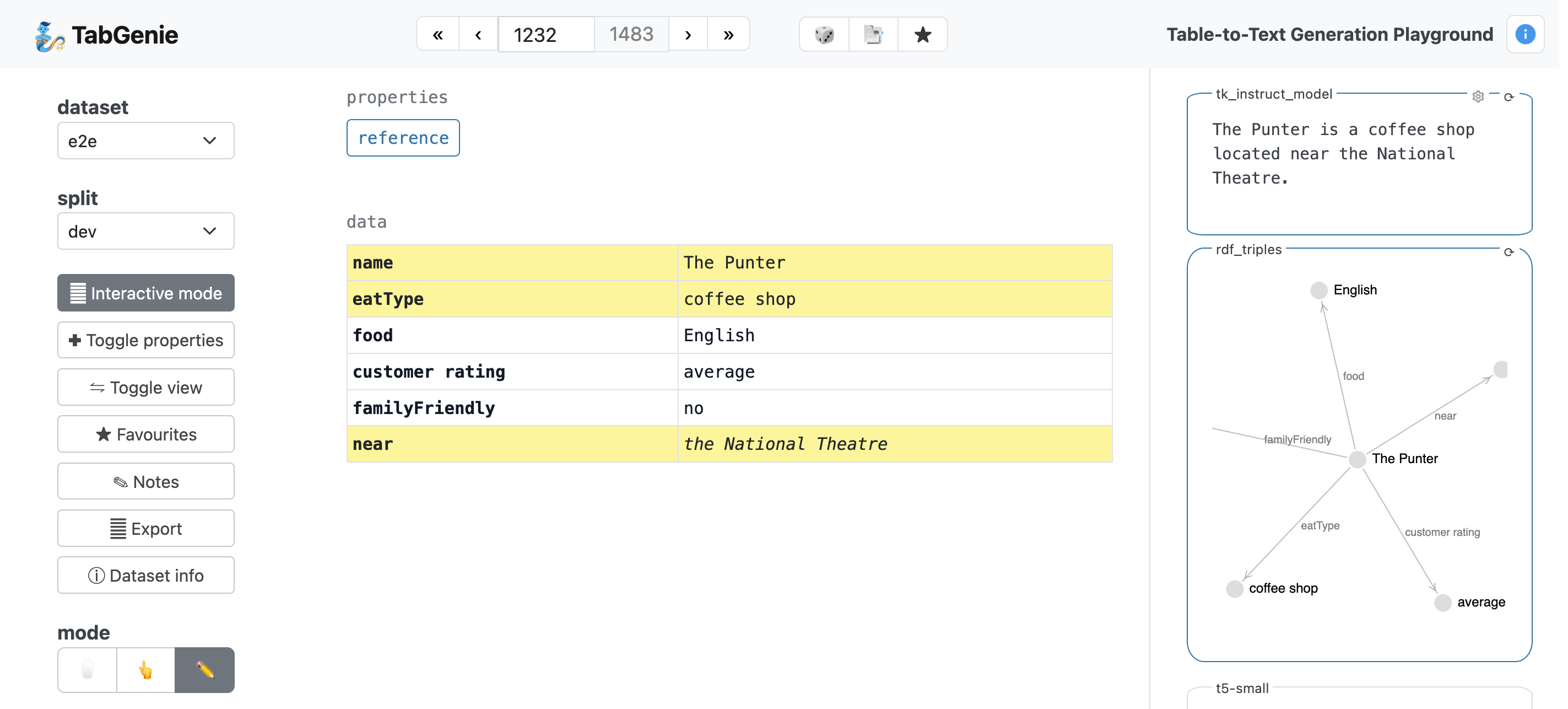}}
  \caption{The interactive mode of the web interface in which the user (1) highlighted specific cells (the cells with the yellow background), (2) edited the input in one of the cells (``\textit{Café Sicilia}'' $\rightarrow$ ``\textit{the National Theatre}''), (3) re-generated the model output (see the top right panel). The figure also shows the graph visualization of the input key-value pairs.}\label{fig:interactive}
\end{figure*}

\begin{figure*}[htb]
  \centering
  \includegraphics[width=\textwidth]{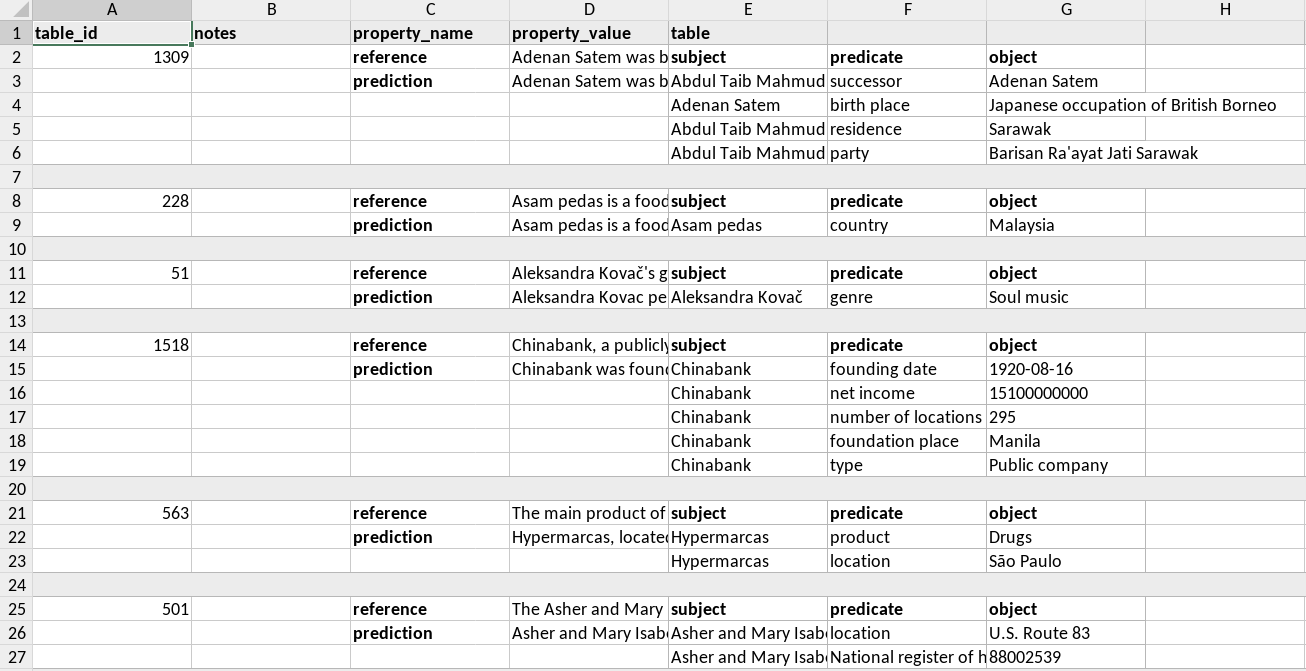}
  \caption{The spreadsheet for manual annotations with a random sample of system outputs exported using \textsc{TabGenie}.}\label{fig:spreadsheet}
\end{figure*}

Figure \ref{fig:interactive} shows the interactive mode in the \textsc{TabGenie} web interface. Figure \ref{fig:spreadsheet} shows the spreadsheet for manual annotations generated using \textsc{TabGenie}.

\end{document}